\documentclass[conference]{IEEEtran}
\usepackage{booktabs}
\usepackage{commath}
\usepackage{algorithm, algorithmic, amsmath}
\usepackage{array}
\usepackage[normalem]{ulem}
\usepackage{siunitx}
\usepackage{adjustbox}
\usepackage{multirow}
\newcolumntype{M}[1]{>{\centering\arraybackslash}m{#1}}
\newcolumntype{N}{@{}m{0pt}@{}}

\usepackage[switch]{lineno}
\usepackage{amssymb}
\usepackage{xcolor}
\usepackage{graphics,graphicx}
\ifCLASSOPTIONcompsoc
  \usepackage[caption=false,font=normalsize,labelfont=sf,textfont=sf]{subfig}
\else
  \usepackage[caption=false,font=footnotesize]{subfig}
\fi

\begin{document}
\title{FLARE: Detection and Mitigation of Concept Drift for Federated Learning based IoT Deployments}
\author{\IEEEauthorblockN{Theo Chow\IEEEauthorrefmark{2}\IEEEauthorrefmark{3},
Usman Raza\IEEEauthorrefmark{4}, Ioannis Mavromatis\IEEEauthorrefmark{2},
Aftab Khan\IEEEauthorrefmark{1}\IEEEauthorrefmark{2},\\
\IEEEauthorblockA{\IEEEauthorrefmark{2}Toshiba Europe Ltd., Bristol Research \& Innovation Laboratory, Bristol, UK\\ 
\IEEEauthorrefmark{3}King's College London, UK\\
\IEEEauthorrefmark{4}Waymap Limited, London, UK
}
Emails: theo.chow@kcl.ac.uk, usman.raza@waymap.org,  \{ioannis.mavromatis, aftab.khan\}@toshiba-bril.com}}

%



\maketitle
\begin{abstract}
Intelligent, large-scale IoT ecosystems have become possible due to recent advancements in sensing technologies, distributed learning, and low-power inference in embedded devices. In traditional cloud-centric approaches, raw data is transmitted to a central server for training and inference purposes. On the other hand, Federated Learning migrates both tasks closer to the edge nodes and endpoints. This allows for a significant reduction in data exchange while preserving the privacy of users. Trained models, though, may under-perform in dynamic environments due to changes in the data distribution, affecting the model's ability to infer accurately; this is referred to as concept drift. Such drift may also be adversarial in nature. Therefore, it is of paramount importance to detect such behaviours promptly. In order to simultaneously reduce communication traffic and maintain the integrity of inference models, we introduce FLARE, a novel lightweight dual-scheduler FL framework that conditionally transfers training data, and deploys models between edge and sensor endpoints based on observing the model's training behaviour and inference statistics, respectively. We show that FLARE can significantly reduce the amount of data exchanged between edge and sensor nodes compared to fixed-interval scheduling methods (over 5x reduction), is easily scalable to larger systems, and can successfully detect concept drift reactively with at least a 16x reduction in latency.
\end{abstract}


%
\IEEEpeerreviewmaketitle

\section{Introduction}
%
%
%
%
Internet of Things (IoT) devices have been widely deployed across various industrial and non-industrial environments to enhance and maintain different services. These include critical applications in healthcare \cite{kodali2015implementation}, manufacturing and product life cycles, warehouse inventory management, etc. \cite{zhong2018internet,tejesh2018warehouse}. 
In the majority of these cases, IoT devices must meet real-time performance and deployment constraints such as low power, small physical size, low manufacturing costs and low installation complexity \cite{capra2019edge}.


In the past, IoT data were processed in a centralised ML architecture. When considering the data exchange cost and the ever-growing number of IoT devices, results in centralised ML becoming prohibitively expensive. Therefore, distributed ML architectures such as the Federated Learning (FL) frameworks \cite{mcmahan_ramage_2017} are now commonly used. Data collected by IoT sensors is sent to Edge devices for training or inference. In an FL setup, multiple edge devices locally train their models and later share them with a central parameter server to be aggregated into a global model. This global model is later sent back to the edge devices for continued learning. In such a setup, the system is able to reap the benefits of models trained from rich data while preserving data privacy. 

In IoT systems, embedded microcontrollers were traditionally used only for sensing purposes such as light, temperature and humidity measurements \cite{sehrawat2019smart}. Akin to advances in edge processing technologies, these embedded devices are becoming increasingly more powerful and capable of running ML inference tasks while still generating and processing the raw data \cite{10060797}. Typically, a pre-trained model is converted into its embedded format and deployed on the resource-constrained embedded sensors. This significantly reduces the data exchange in the entire system and enhances the system's scalability and efficiency while reducing the cost.

However, real-world systems, being highly dynamic environments, introduce significant challenges in the pre-trained ML models. ML models, as the underlying relationship between the input (e.g., sensing data) and output (target) variables changes over time, become outdated and their performance drops. This behaviour is called concept drift~\cite{concept_drift} and can occur for several reasons, e.g., long-term climate changes, short-term sensor drift, etc. Concept drift can also result from adversarial attacks, such as data poisoning attacks which can be even more detrimental for FL deployments. Even if one node is attacked in an FL setup and its data is poisoned, the attack can disperse across all other clients as all models are aggregated to a single global one.


Concept drift is mitigated against with frequent retraining of the model with recent data and non-poisoned data. In the IoT context, even though embedded devices are capable of performing inference tasks, training is usually conducted on the edge. Thus, there is always a tradeoff between the data exchanged and the expected model performance that should be considered, particularly in resource-constrained environments. Considering all the above, we present Federated LeArning with REactive monitoring of concept drift (FLARE), a novel scheduling method of ensuring sustained model performance while minimising the communication overhead in a cloud-edge-endpoint continuum. More specifically, our contributions include: 
\begin{itemize}
    \item A scheduling algorithm deployed within the training node (e.g., an FL client) for assessing the model's status, and deploying it on a sensor/endpoint node for inference, when it is in an optimal state.
 	\item Another scheduling algorithm at the sensor to observe model status using statistical testing during deployment at the sensor node where inference is performed. Our proposed approach for maintaining the quality of the deployed model does not rely on the ground truth and solely uses model confidences at inference.
 	\item FLARE is extensively evaluated using the MNIST-corrupted dataset by exposing it to various drift levels for three different types of corruptions.
 	\item We perform evaluations in both small- and large-scale Federated Learning-based deployments in which various KPIs are compared against the benchmark schemes.
 \end{itemize}

\begin{figure}[t]
    \centering
    \includegraphics[width=0.8\columnwidth]{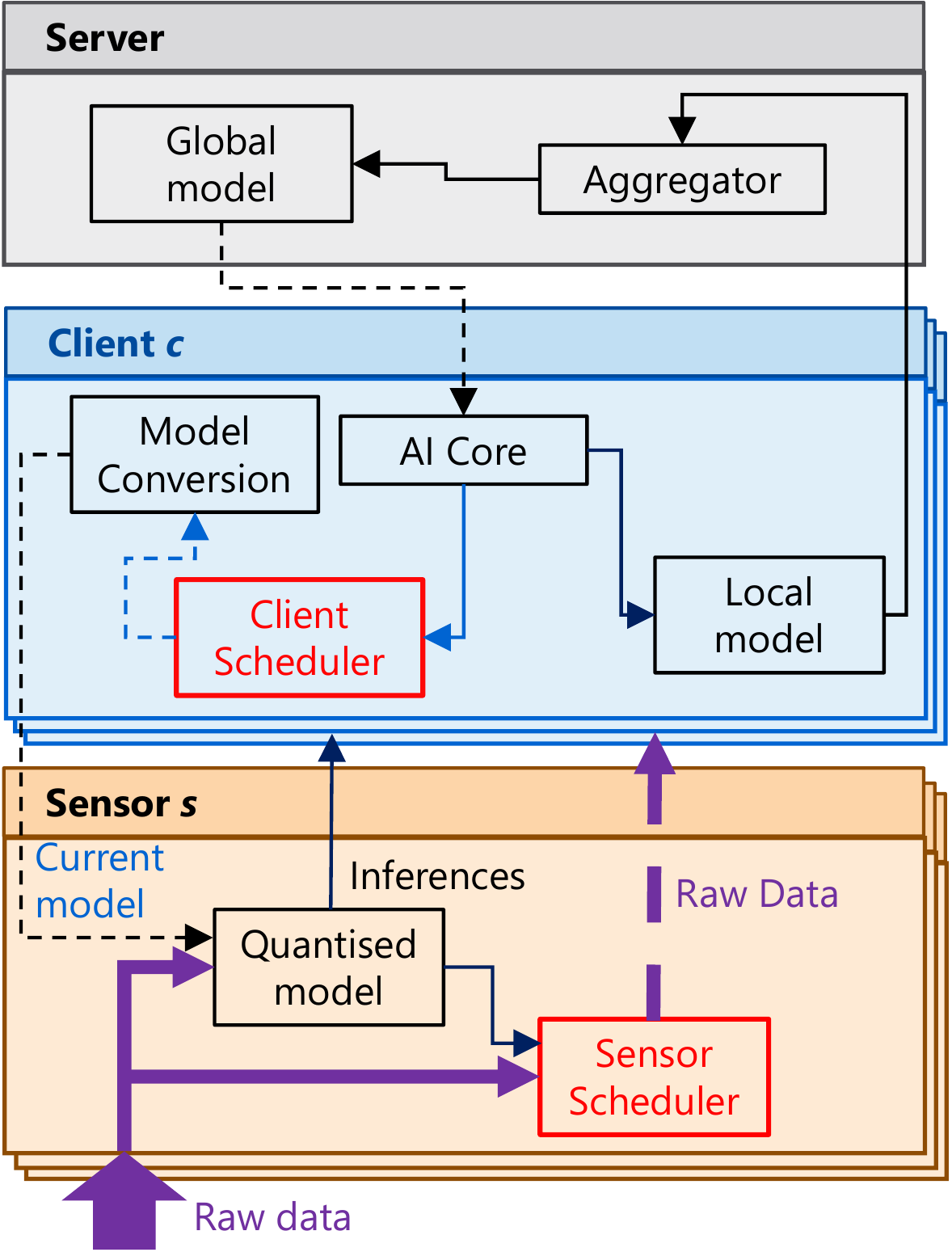}
    \caption{Overview of FLARE showing data communication links among three nodes (server, client, and sensor). Solid lines indicate constant communication whereas dashed lines indicate conditional communication.}
    \label{fig:Blockdiagram}
\end{figure}

\section{Related work}
\label{sec:CD}
ML models typically require training with large amounts of data before they can be deployed for inference. Training data would provide a good representation of the data collected at inference time. However, when the environment is dynamic, the data distribution changes over time, leading to a deterioration in the trained model's accuracy. This change in data distribution during inference time is known as concept drift and is especially detrimental to long-term ML deployments if poorly maintained \cite{gama2014survey}. There are several types of concept drift, and previous work in the literature has included different methods in categorising them \cite{webb2016characterizing}. Concept drift problems are often addressed through statistical methods, such as comparing a statistic representing the similarity between two data sets \cite{dries2009adaptive,de2018concept}.

Concept drift on centralised continuous architectures can affect the long-term accuracy of one single model. However, concept drift on distributed decentralised continuous learning systems will affect the entire system, including every edge node. In the case of FL, a change in distribution at one of the sensors (data poisoning) would directly impact other local models. Therefore, it is imperative to detect concept drift for large-scale FL deployments. Previous work has explored the effects of concept drift for FL and continual settings, however, they require specific conditions. For example, CDA-FedAvg detects drift during training and thus requires the availability of drifted labels which runs the risk of missing drift\cite{casado2020concept}. 
Hassan Mehmood et al. presented a method of detecting concept drift using time series data but did not consider the model drift itself \cite{mehmood2021concept}. 
Furthermore, the detection methods rely on absolute confidence values or differences between previous and current confidence values. DNNs often provide highly confident predictions that can be inaccurate and unreliable \cite{nguyen2015deep}. Our proposed methods (detailed below) rely on the change in the distribution of the confidence values under different conditions, providing a more reliable and efficient approach for resource-constrained devices.

\section{System Overview}
This paper, based on~\cite{theo2023system}, proposes FLARE, which incorporates two scheduling subsystems for deployment on training nodes (e.g., a cloud server in centralised or clients in federated systems) and sensors (low power, computationally constrained embedded devices).
The proposed solution, therefore, consists of two schedulers, one placed at the client and another at the sensor. This work concentrates on deploying this approach in a federated learning setting. These two subsystems can be implemented and deployed independently, but since both methods complement each other, we deploy them in tandem to effectively optimise overall data communication. Figure \ref{fig:Blockdiagram} shows the architectural diagram of the entire system when used in an FL environment. It consists of a server where a global model is initialised and shared among clients for training. Clients contain processing units (such as GPUs) to train ML models (represented as their AI core at the edge) with their local datasets and produce individual local models. These local models are continuously shared with the server for aggregation. 

Our first proposed scheduler observes model training and assesses when the model is ready to be deployed for inference. As illustrated in Figure \ref{fig:Blockdiagram}, the client scheduler effectively decides a suitable deployment time, after which models are converted to embedded/quantised format ready for inference. During inference, the model's confidences are observed with the second scheduler that decides whether the model has drifted. In the case of drift detection, a mitigation strategy is shared; in this case, data is shared with the client for training the model with the latest data. Details of both the proposed scheduling schemes are provided below.

\section{Methodology}
The proposed environment consists of three separate nodes; Sensor $s$, Client $c$ and a server. 
With the introduction of the two scheduling subsystems, we can restrict the data communication between the three nodes for efficient data communication with minimal sacrifice in inference accuracy.
%
{
Table \ref{tab:notations} lists all the notations used in this paper.}

\begin{table}[t]
    \caption{Table of notations.}
    \centering
    {
    \begin{tabular}{lp{0.78\columnwidth}}
    \toprule
    Notation & Description \\
    \midrule
    $w$ & Time window 
    \\
	$\Delta$ & Absolute loss difference values \\
	$\alpha$ & Model instability coefficient \\
	$\beta$ & Model stability coefficient \\	
	$\phi$ & Sensor test data distribution threshold \\
	$\theta_{ks}$ & Kolmogorov-Smirnov test statistic\\
	$\sigma_w$ & Current Standard deviation of the absolute loss differences\\
    $\sigma_s$ & Stable Standard deviation of the absolute loss differences\\
	\bottomrule
    \end{tabular}}
    \label{tab:notations}
\end{table}



\paragraph{\textbf{Client}}
FL systems \cite{mcmahan_ramage_2017}, as introduced above, 
rely on a training node called the \textit{client}.
At the client, local models are trained after receiving an initial global model from the \textit{server}.
Our proposed scheduler system runs within the client to evaluate model stability during training.
This is achieved using a subset of the training data to validate the model's stability.
The losses of the training and validation sets can be calculated using the local model trained on the client.
Model stability is determined by comparing the standard deviation of the absolute loss differences using the two sets and the mean of the absolute loss difference. 

During a period of instability, two possible actions can be taken, \emph{i)} if the model becomes stable, it is converted into an embedded format and sent to the sensor for deployment (this conversion step is only required for sensor nodes where only embedded inferences are supported), and \emph{ii)} if the model remains unstable, the model will continue training with the existing training data at the client.
Formally, in each time window $w$, an array of validation loss $\lambda^{v}$, and training loss $\lambda^{tr}$, are calculated. These losses then are used to calculate the standard deviation $\sigma_w$ via the absolute loss differences, $\Delta$. 
\begin{equation}
	\Delta = \mid \lambda^{tr} - \lambda^{v} \mid
\end{equation}
where $\lambda_n^{tr}$ and $\lambda_n^{v}$ represent the training and validation losses of a given sample in the time window. 
These absolute loss differences are then used to calculate the standard deviation. 
\begin{equation}
	\sigma_{w} = \sqrt{\frac{\sum_n^w (\Delta - \mu)}{w-1}}
\end{equation}
By using the standard deviation in the current time window $\sigma_w$ against the previous stable standard deviation value $\sigma_s$ modified by the model's stability coefficients $\alpha$ and $\beta$, we can assess the model's stability. 
Firstly the model is marked as unstable if: the following condition is true:
\begin{equation}
    \sigma_w > \sigma_s \times \alpha,
\end{equation}
where $\sigma_w$ represents the standard deviation in the current time window, $\sigma_s$ represents the previous stable standard deviation value, and $\alpha$ is the model's instability coefficient. During this phase, model training continues until stability is achieved.
%
%
The model is converted to an embedded device format, sent to the sensor, and is marked as stable if it was previously unstable \textit{and} the following condition is true:
    $\sigma_w < \sigma_s  \times ( 1 + \beta )$
where $\beta$ represents the model stability coefficient.
Since model training is a stochastic process, the previous stable standard deviation $\sigma_s$ will change whenever the following condition is true:
\begin{equation}
    \sigma_w < \sigma_s  \times ( 1 - \beta ) 
\end{equation}



    
    

\begin{algorithm}[tp]
\begin{algorithmic}[1]
\caption{Client scheduler subsystem}
\label{alg:WorkerAlg}

\REQUIRE{CalculateLoss(), ConvertModel(), DeployModel(), StandardDeviation(),LossWindow(), $ValD$, $TestD$}
\STATE $ unstable \leftarrow $ False \\
\STATE $\sigma_s$ $ \leftarrow $ 0 \\
\STATE val\_loss  $ \leftarrow $  [ ] \\
\STATE test\_loss  $ \leftarrow $  [ ] \\
\STATE loss\_difference  $ \leftarrow $  [ ] \\

    \LOOP
        \STATE val\_loss $ \leftarrow $ LossWindow(local\_model,$ $ValD$ $, $w$) 
        \STATE test\_loss $ \leftarrow $ LossWindow(local\_model,$ $TestD$ $, $w$) 
        \STATE loss\_difference $ \leftarrow $ $\abs{ \textrm{test\_loss} - \textrm{val\_loss} }$
        \STATE $\sigma_w$ $ \leftarrow $ StandardDeviation(loss\_difference)
    	\IF{ $\sigma_w > \sigma_s \times \alpha$}
    		\STATE $unstable \leftarrow $ True
    
    	\ELSIF{ $\sigma_w < \sigma_s \times (1 - \beta$)}
    		\STATE $\sigma_s$ $ \leftarrow $ $\sigma_w$
    
    	\ELSIF{ $\sigma_w < \sigma_s \times (1 + \beta$) and $ unstable = $ True}
    		\STATE $ unstable  \leftarrow $ False
    		\STATE embedded\_model $ \leftarrow  $ ConvertModel(local\_model)
    		\STATE DeployModel(embedded\_model)
    	\ENDIF
    \ENDLOOP
\end{algorithmic}
\end{algorithm}


In a multi-class classifier that contains $C>2$ classes, input data samples $x_i$ are used to produce a class prediction $y_i$ and a confidence score $p_i$. The network logits $z_i$ can be used to calculate both $y_i$ and $p_i$, typically using the softmax function represented here by $g$:
\begin{equation}
        g(z_i)^c = \frac{\exp(z_i^c)}{\sum_{j=1}^C \exp(z_i^j)}, p_i = 
        \max g(z_i)^K
\end{equation}
This list of confidence scores is utilised at the sensor and will be explained in detail in  the following section.


\paragraph{\textbf{Sensor}}
In order to observe model quality at the sensor node where raw data is collected, the method proposed compares the confidence values generated using the confidence validation set and the test sets. This is done using the Kolmogorov-Smirnov (KS) test \cite{massey1951kolmogorov} that compares the cumulative distribution function (CDF) of the received confidence values with the CDF of the confidence values generated from the test set. 
By evaluating the similarity of the client test confidences and sensor test confidences, two possible decisions can be made, \emph{i)} If the similarity is low, indicating there is a change in the distribution of the sensor test set (for example, due to the addition of noise), the sensor then sends new raw data to the client for further training (in supervised learning, this data would also require labelling), and \emph{ii)} If the similarity is high, indicating the deployed model is still effective for the current data set, the sensor can continue inference without transferring any new data to the client.

The value of the KS test ranges between 0-1, with 0 being high similarity and 1 being low similarity. Suppose there is a change in data distribution. In that case, the value of the KS test will increase, indicating low similarity between the CDFs of the client test confidence values and the sensor test confidence values. Conversely, when the model improves, the KS value will fall to reflect the high similarity between the two CDFs. We detect this change by evaluating if the current KS value is increased by $\phi$ from the previous KS value. 

\begin{figure}[t]
\centering
\subfloat[Zigzag]{\includegraphics[width=1.1in]{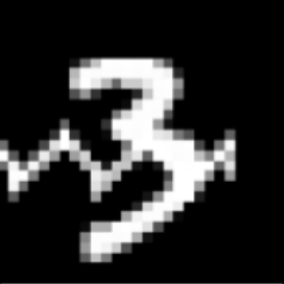}%
\label{fig:Zigzag}}
\hfil
\subfloat[Canny Edges]{\includegraphics[width=1.1in]{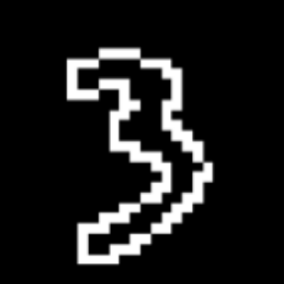}%
\label{fig:CannyEdges}}
\hfil
\subfloat[Glass Blur]{\includegraphics[width=1.1in]{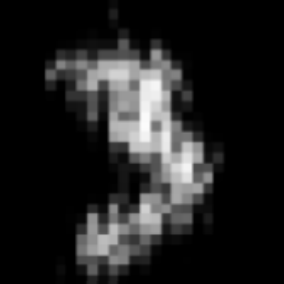}%
\label{fig:GlassBlur}}
\caption{Samples of corrupted MNIST images using three corruption methods.}
\label{fig:CorruptedMnist}
\end{figure}

\begin{figure*}[t]
\centering
\subfloat[Accuracy over time in three different scheduling schemes.]{\includegraphics[width=1\columnwidth]{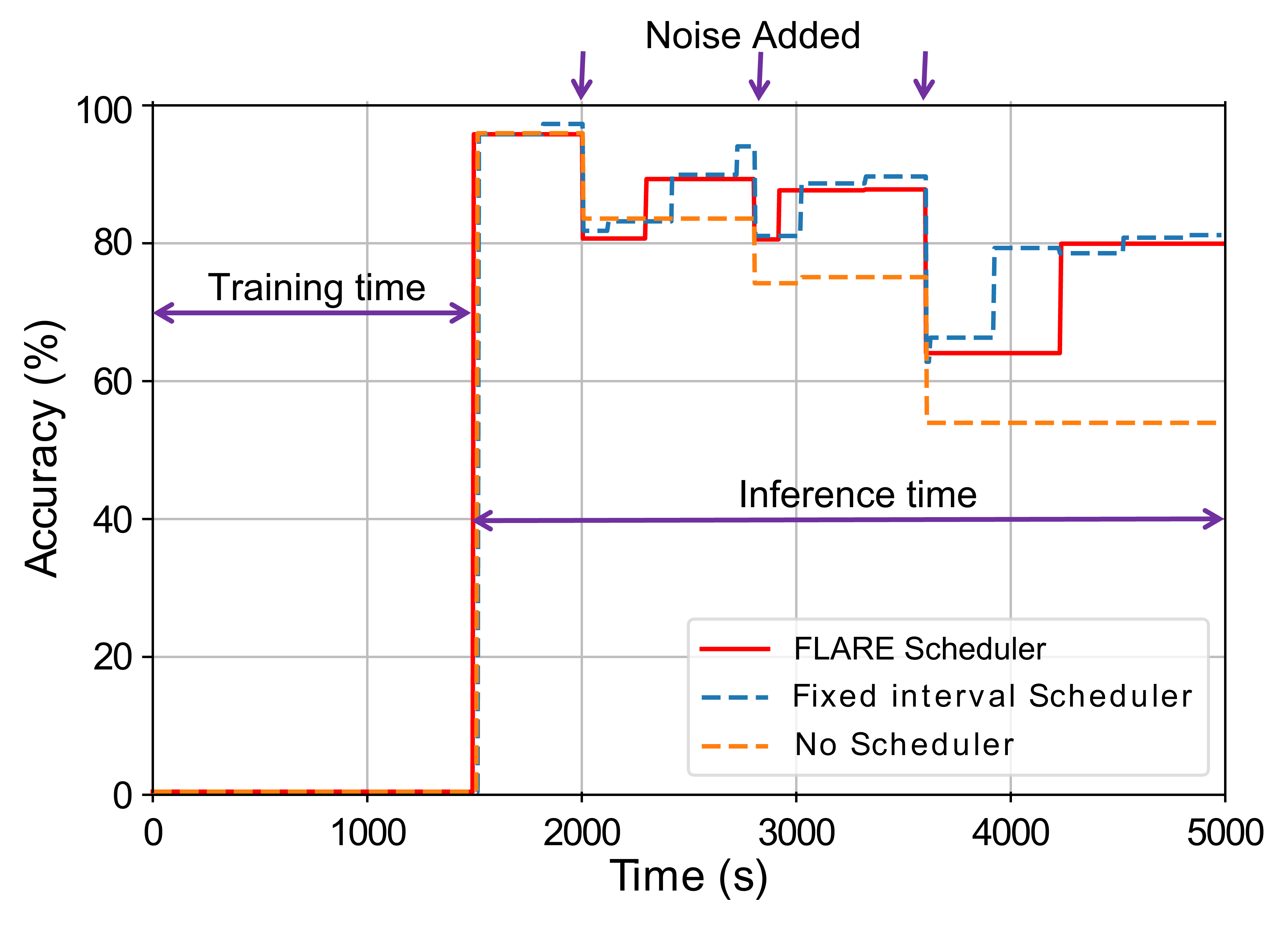}%
\label{fig:AccuracySmallScaleFL}}
\hfil
\subfloat[Overall data communicated for FLARE and fixed interval scheduler.]{
\includegraphics[width=1\columnwidth]{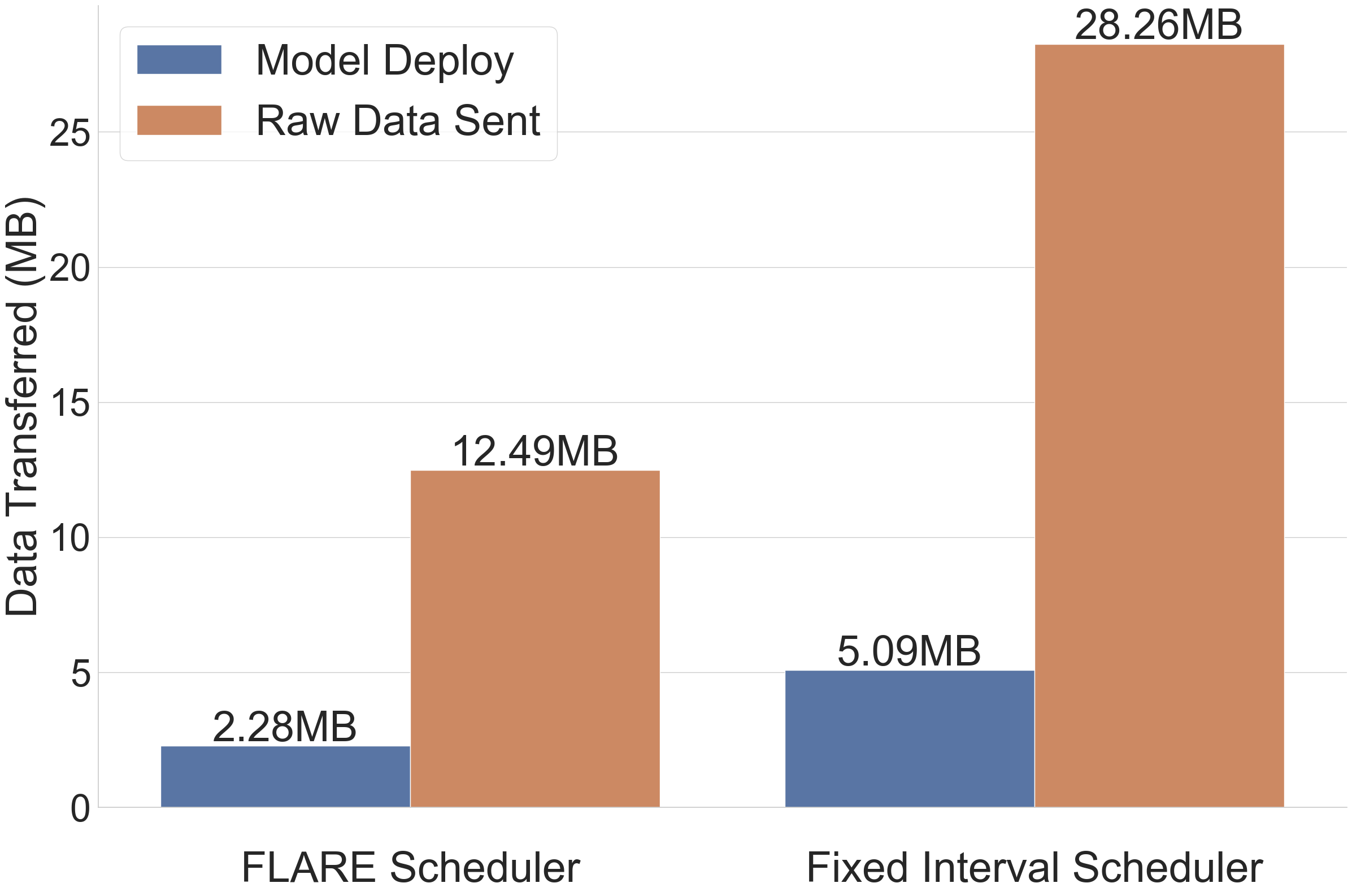}
\label{fig:DataSmallScaleFL}}
\caption{Comparison results for the preliminary FL experiment. Drift is added at \SI{2000}{\second}, \SI{2800}{\second} and \SI{3600}{\second}.}
\label{fig:SmallScaleResults}
\end{figure*}

\section{Experiments}
We benchmark FLARE in two different scenarios. For both experiments, we assume: a) deploying a model on the sensor, and b) transferring the data from sensors to the clients; both at fixed intervals. We begin with a preliminary study comprised of one sensor and one client, and compare it against a fixed scheduler and a setup with no scheduling method. We later investigate a more real-world-like scenario (4 clients, 32 sensors) comparing FLARE against two fixed interval schedulers, i.e., high- and low-frequency schemes. Our Key Performance Indicators (KPIs) are classification accuracy, communication volume, and drift detection latency.

\subsection{Dataset Description}
ML models deployed in production environments experience different types of drifts (see Section \ref{sec:CD}). In our experiments, we primarily focus on \textit{abrupt} drift changes. MNIST Corrupted~\cite{mu2019mnist} dataset is well suited for such drifts containing 15 types of corruptions applied on handwritten digits. We chose three, i.e.,  \textit{Zigzag}, \textit{Canny edges}, and \textit{Glass blur} (Figure~\ref{fig:CorruptedMnist}) and introduce them at the sensor's data with fixed intervals after the initial deployment (mainly once the model is trained for a fixed initialisation period).
For drift mitigation, we assumed these changes in data are benign in nature and as such are incorporated as new data for training within the FL system.
All the different sub-datasets in the client and sensor are set to a fixed size to keep the data distribution consistent, and our evaluation focused on the system's ability to detect and mitigate concept drift.

\subsection{Model architecture}
Deep CNNs are known to perform well in image classification problems. However, due to the hardware constraints on embedded devices, a very deep CNN model is not typically optimal. As such, we opted for a basic CNN architecture (with two convolutional layers with max-pooling followed by two Dense layers; all with Relu activation function and softmax in the final layer). That can easily be optimised for embedded devices whilst still retaining high accuracy. We used a Gradient descent optimizer with a learning rate of 0.1 (fixed across all of the experiments).
%
 %

\subsection{Parameter Optimisation}
Selecting different values for $\alpha$ (model's instability coefficient at the edge), $\beta$ (model's stability coefficient at the edge) and $\phi$ (sensor test data distribution threshold at the sensor)
directly impacts the frequency of communications:
\begin{align}
    \alpha &\in \mathbb{R} \enspace | \enspace \alpha \geq 0 \\
    \beta &\in \mathbb{R} \enspace | \enspace 0 \leq \beta \leq \alpha \\
    \phi &\in \mathbb{R} \enspace | \enspace 0 \leq \phi \leq 1
\end{align}
where a larger value of $\alpha$ decreases the sensitivity to concept drift detection and reduces communications at the client. 
On the other hand, higher $\beta$ will decrease the sensitivity to concept drift detection and increase communications. 
Lastly, $\phi$ has a similar effect as $\alpha$ and decreases sensitivity to concept drift detection with a higher value at the sensor. 
In this work, we use: $\alpha$ = 8, $\beta$ = 0.3, $\phi$ = 0.2 and $w$ = 10 (time window used for calculating the losses). All values were empirically picked utilising the validation set. 
These parameters can also be automatically determined and adjusted based on the available bandwidth or performance requirements. However, in this paper, we used static values in order to keep the experimental evaluation focused on the ability of the proposed approach in detecting and mitigating concept drift in FL deployments.


\section{Results}
For both experiments, we introduce drift at pre-configured fixed intervals but after allowing sufficient time for the models to train; this also reflects realistic deployment scenarios where ML model inferences are collected only after models are sufficiently trained. In our experiments, different corrupted images are added to the inference set after this initial training period and while the model is deployed on the sensors performing inference.

\begin{figure*}[t]
     \centering
     \begingroup
     \captionsetup[subfigure]{margin={1cm,0cm}} 
     \subfloat[High-freq. Fixed interval Scheduler.]{\includegraphics[height=4in]{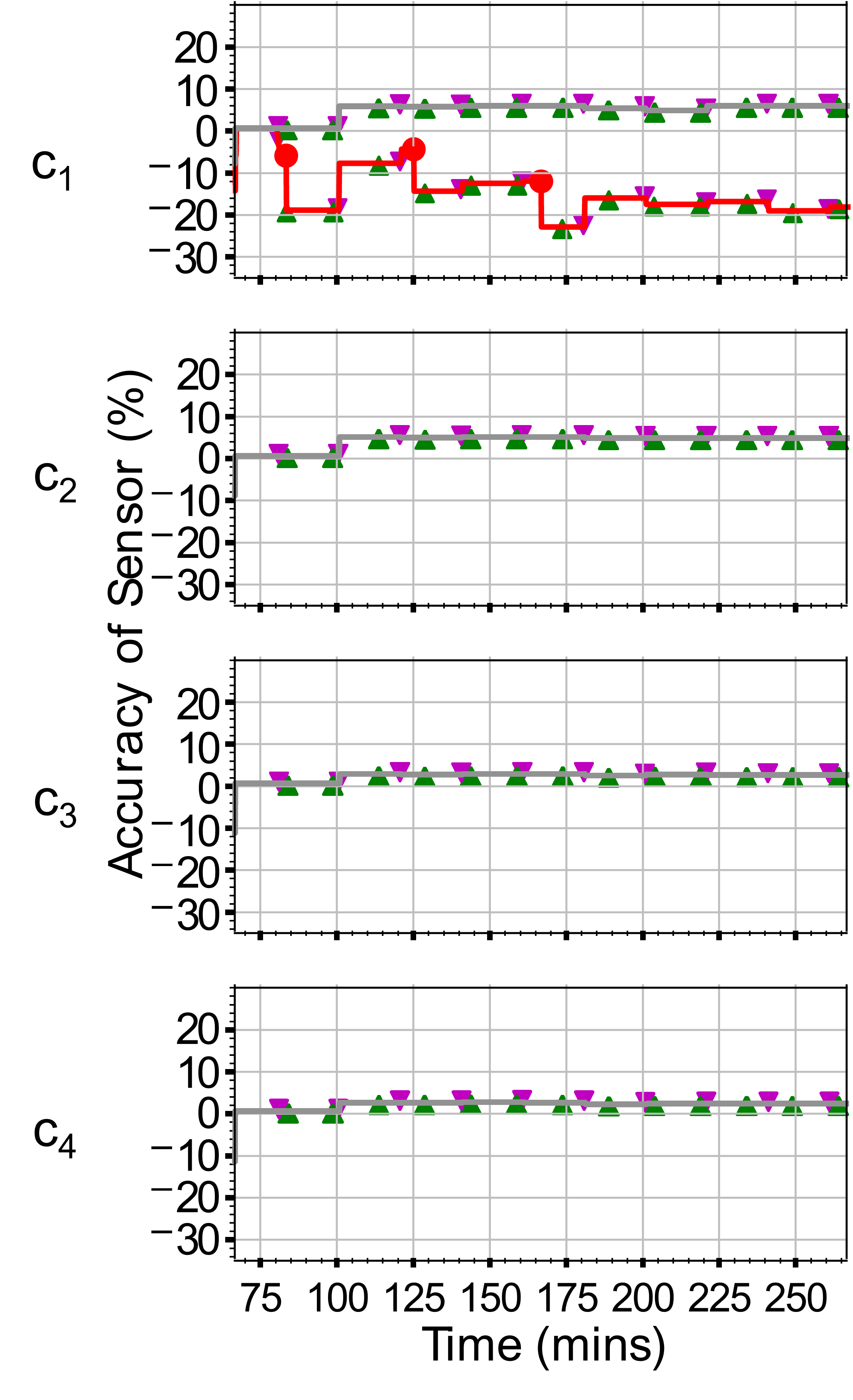}}
     \endgroup
     \qquad
     \begingroup
     \captionsetup[subfigure]{margin=0cm} 
     \subfloat[Low-freq. Fixed interval Scheduler.]{\includegraphics[height=4in]{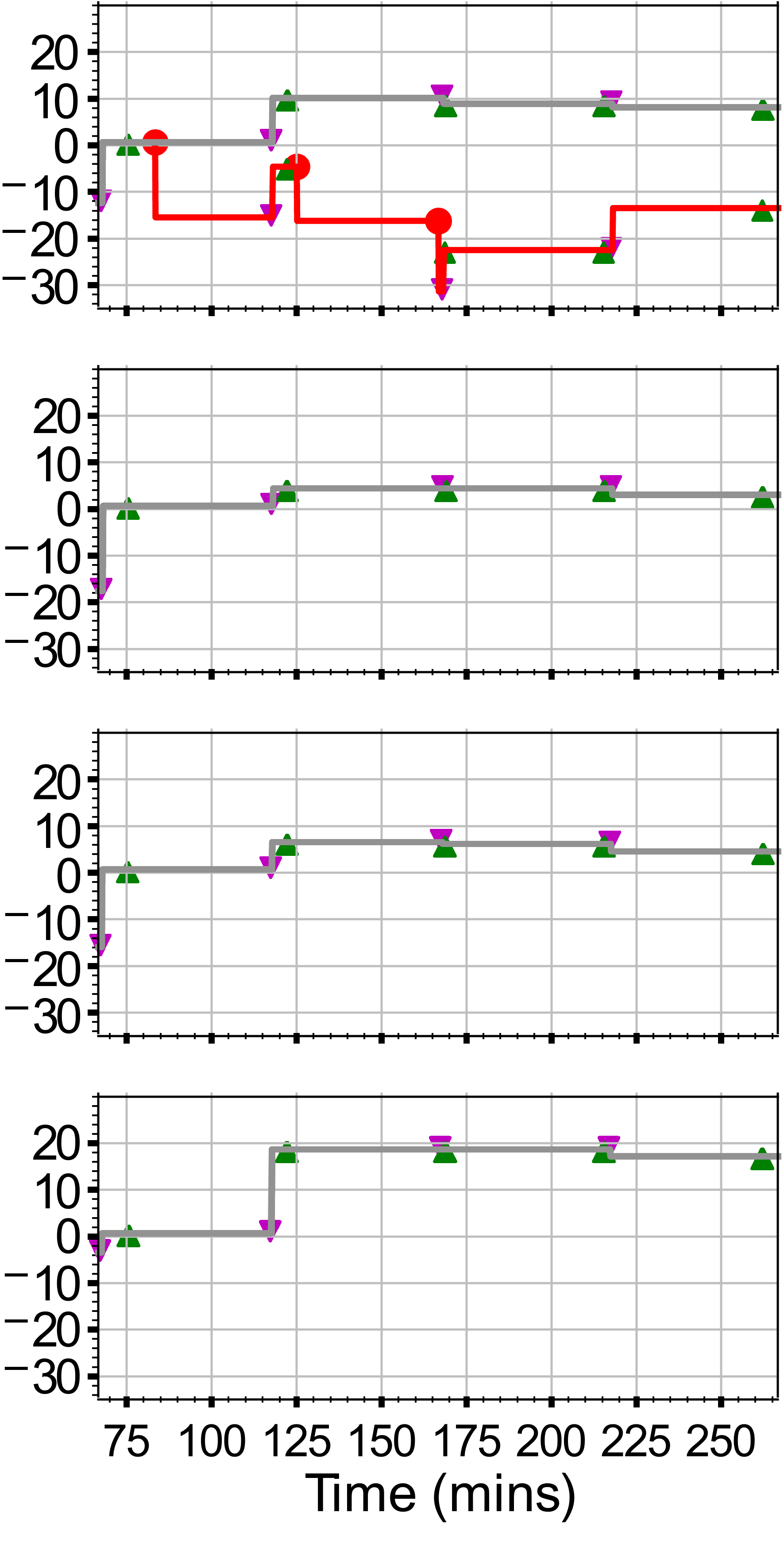}}
     \qquad
     \subfloat[FLARE Scheduler.]{\includegraphics[height=4in]{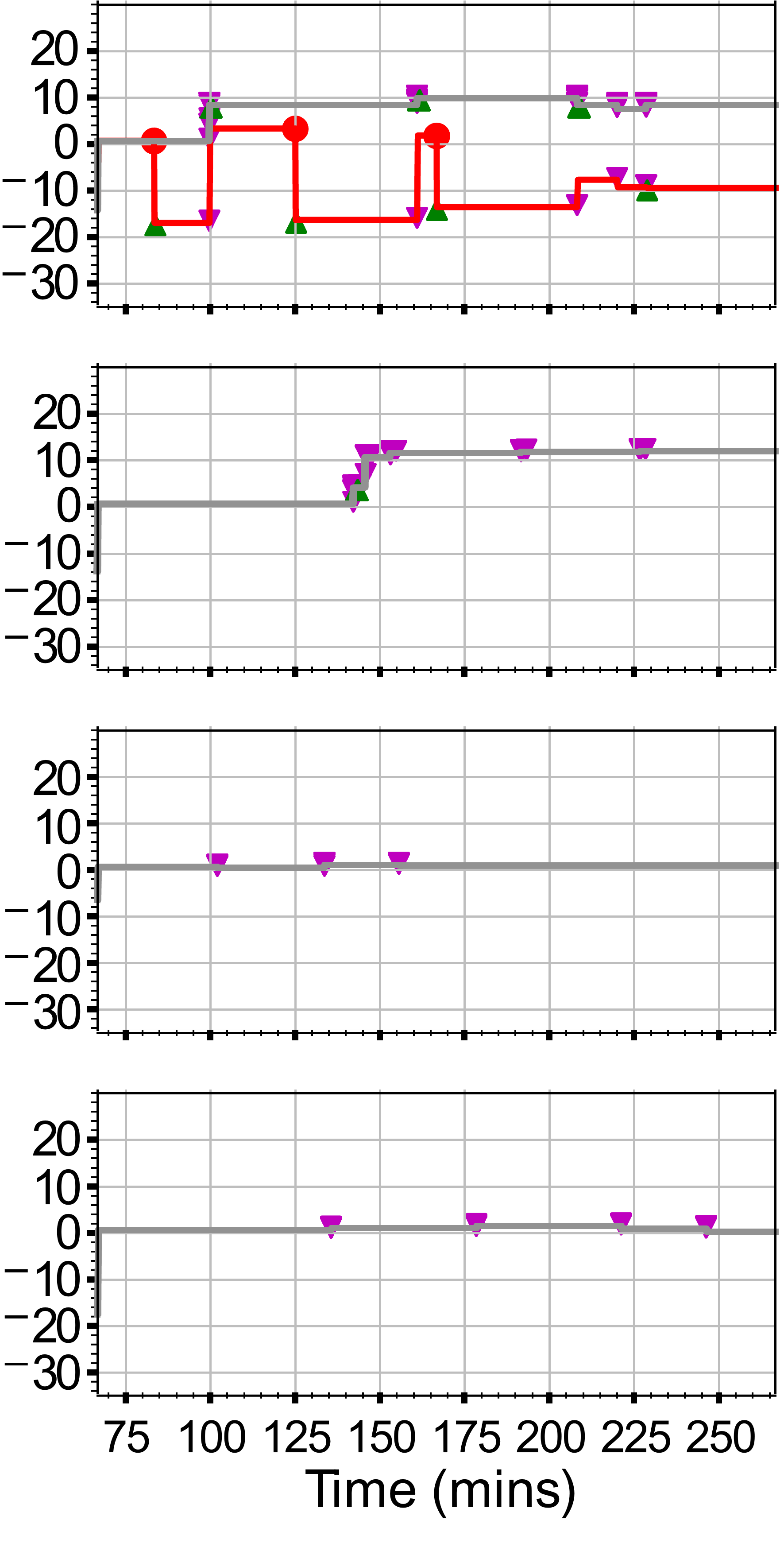}}    
     \endgroup
    \caption{Comparison of accuracy for real-world-like FL deployment using FLARE against the baseline approaches. Magenta triangle $\begingroup\color[rgb]{0.76, 0, 0.47}\blacktriangledown\endgroup$ indicates a model being deployed to the sensor (downlink). Green triangle $\begingroup\color[rgb]{0.082, 0.69, 0.101}\blacktriangle\endgroup$ indicates new raw data being communicated to the client for training (uplink). Red dot $\begingroup\color{red}\bullet\endgroup$ indicates new noisy data introduced at the target sensor endpoint. Red lines $\begingroup\color{red}\mathbf{-}\endgroup$ represent the sensors where the noise was introduced whereas grey lines $\begingroup\color{gray}-\endgroup$ are the sensors not directly affected by noise. The high-frequency fixed scheduler constantly receives new data from the sensor and similarly constantly deploys models resulting in greater instances of uplink and downlink traffic, and vice versa for the low-frequency scheduler.}
    \label{table:AccResult}
\end{figure*}

\subsection{Preliminary FL Experiment}
During our preliminary experiment, we compared our proposed methods with two alternative schemes, i.e., fixed interval and no scheduling. For the first \SI{1500}{\second}, we utilise the data to train the model allowing sufficient time for the model to be pre-trained. At \SI{1500}{s} the trained model is deployed to the sensor. For the fixed interval scheduling experiment, we deploy a new model at fixed intervals of \SI{300}{\second} whereas raw data is sent to the client every \SI{350}{\second}. For the experiment without scheduling, no model is deployed except the first one at \SI{1500}{\second} and no data is sent back to the client thereafter. New drift is added \SI{500}{\second} seconds after the initial deployment of the model and \SI{800}{\second} subsequently.

Figure~\ref{fig:AccuracySmallScaleFL} shows the accuracy perceived at the sensor when different scheduling schemes are used. Our results show that the accuracy using FLARE recovers well after every new introduction of corrupted images. This indicates the system is able to detect, re-train, and re-deploy without manual intervention. 
In the case of no scheduler, significant performance deterioration of the model is observed. 
When compared to the accuracy perceived at the sensor with a fixed interval scheme, FLARE scheduling closely matches it but does not follow it completely. This is due to the higher frequency of communications, as seen in Figure~\ref{fig:DataSmallScaleFL}. In this, we essentially demonstrate that it is not required to constantly communicate data between the client and the sensor. Instead, conditional communication can significantly reduce the total data transferred. It is important to note that sending raw data is considerably more costly than re-deploying a model. Therefore, simply limiting the transfer of raw data already drastically reduces the total data transferred.


\begin{figure*}[t]
\centering
\subfloat[Data transferred between client 1 and affected sensor.]{\includegraphics[width=0.98\columnwidth]{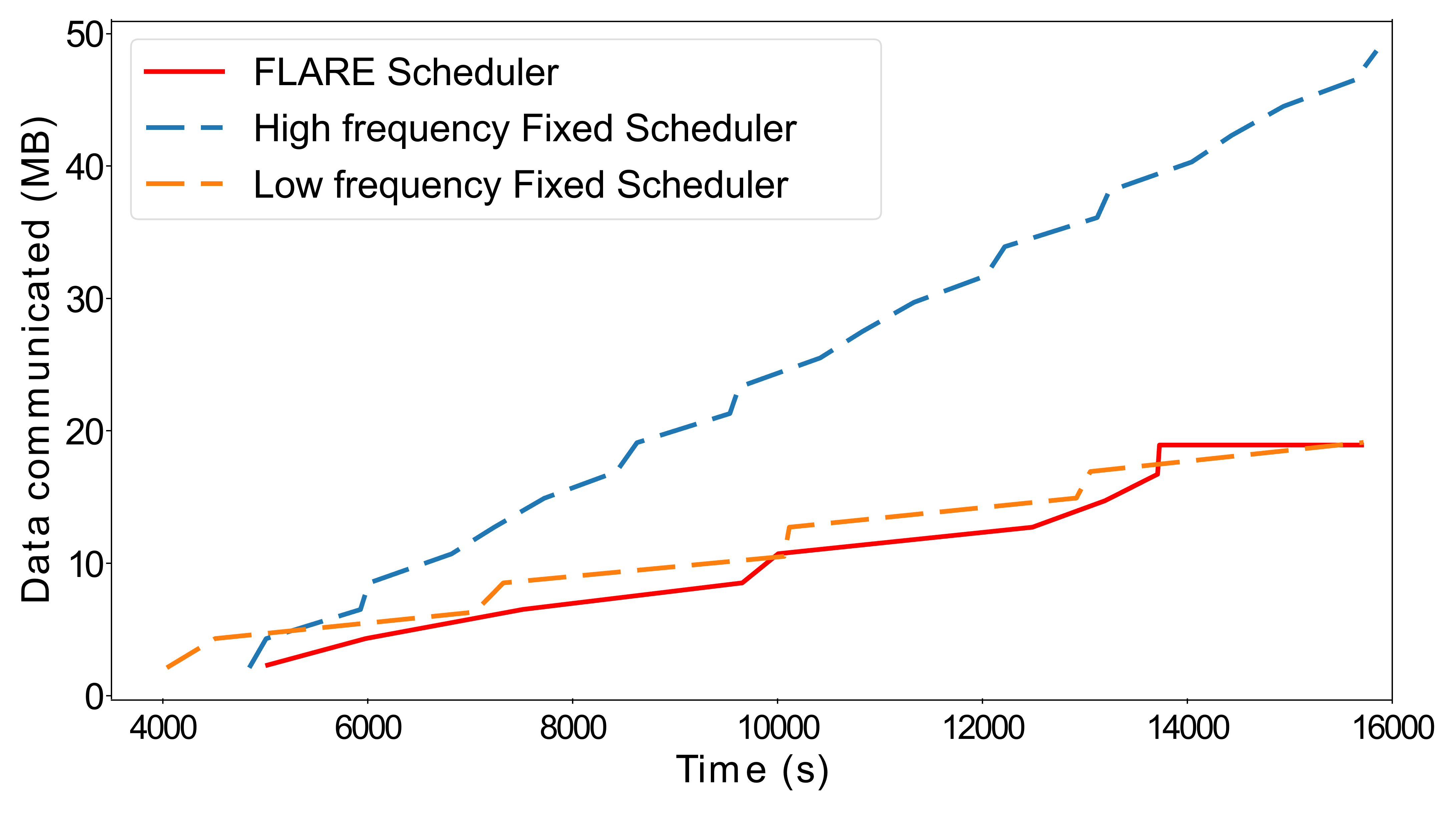}%
\label{fig:IsolatedData}}
\hfil
\subfloat[Data exchanged in the entire FL system.]{\includegraphics[width=1\columnwidth]{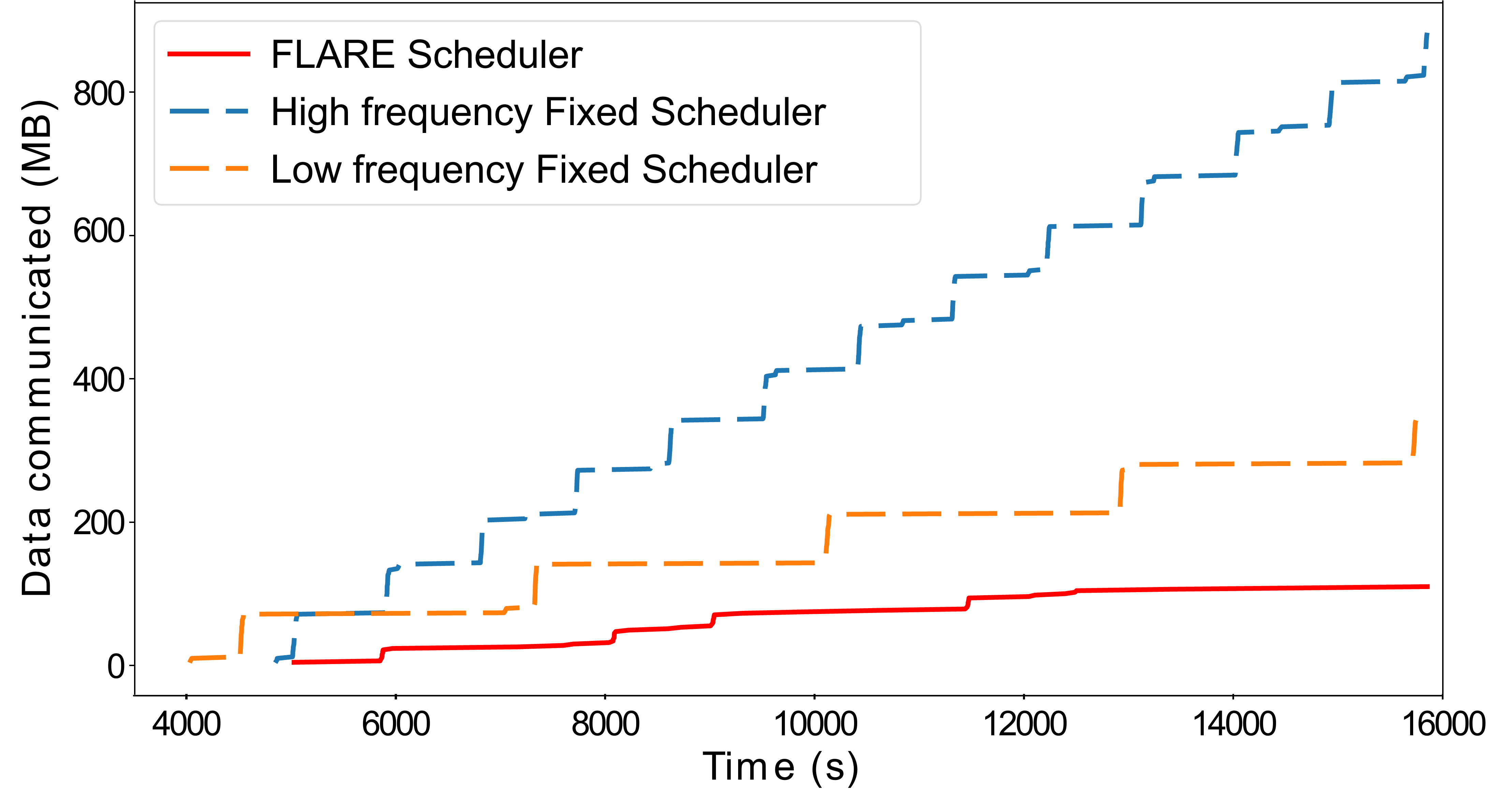}%
\label{fig:FullData}}
\caption{Cumulative data transmission comparison for a real-world-like FL deployment, highlighting the lack of scalability in fixed interval scheduler schemes.}
\label{fig:DataTransferred}
\end{figure*}

\subsection{Real-world FL Experiment} 
For the larger real-world-like experiment, we use a multi-sensor, multi-client environment (with four clients connected to 8 sensors each). For this experiment, we introduce corrupted images to one of the 32 sensors demonstrating a realistic scenario (e.g., a faulty sensor or a malicious action to one of the devices). The rest of the sub-datasets used on the other sensors are kept intact. We extend the experiments for all clients to pre-train until \SI{4000}{\second} (this allows sufficient pre-training prior to initial deployment). Corrupted MNIST images are introduced to the given sensor \SI{1000}{\second} after initial deployment and \SI{2500}{\second} subsequently. 

In this setup, we compare FLARE against two fixed interval schedulers with different intervals. We fix our high-frequency interval scheduler to deploy every \SI{1200}{\second} and send new data every \SI{900}{\second} and our low-frequency interval scheduler to deploy every \SI{3000}{\second} and send new data every \SI{2800}{\second}. Due to the randomness of ML training, we normalise the inference accuracy to the initial deployment. This allows for a clear view of the effect of the drift and recovery of the sensor. 

For FLARE, we observe a consistent accuracy with no more than 18\% maximum drop. This is comparable to the 17.5\% drop in a high-frequency fixed interval scheduler but much lower than the 32.5\% seen in the low-frequency setup. Both high and low fixed schedulers are able to further recover to a 12.5\% and 14.5\% difference in accuracy after several more deployments, but FLARE recovers to a final accuracy difference of 10.2\%. 
Interestingly, as shown in Figure ~\ref{table:AccResult}, the drift effect in the sensors of client 1 does not carry to the other sensors in other clients. Small fluctuations in accuracy are likely due to the FL training at the clients.


\begin{table}[t]
\caption{Comparison of detection latency for a real-world-like FL deployment using FLARE against the baseline approaches.}
\centering
\resizebox{1\columnwidth}{!}{%
\begin{tabular}{lcccc}
\toprule
Scheduling scheme & N1 & N2 & N3 & Average\\
\midrule
High-frequency Fixed interval & \SI{7}{\second} & \SI{223}{\second} & \SI{415}{\second} & \SI{215}{\second}\\
Low-frequency Fixed interval & \SI{2324}{\second} & \SI{2615}{\second} & \SI{113}{\second} & \SI{1684}{\second}\\
FLARE & \SI{22}{\second} & \SI{15}{\second} & \SI{3}{\second} & \SI{13}{\second}\\
\bottomrule
\end{tabular}}
\label{table:LatencyResult}
\end{table}
\subsection{Assessing the Drift detection latency} 
To assess the drift detection latency, we also compared the average time a sensor takes to send raw data to the client after drift is added (for the first time). We took the average from the three different types of drift (Figure~\ref{fig:CorruptedMnist}) added to determine the final latency of a given scheduling system (see Table~\ref{table:LatencyResult}).
%
%
FLARE outperforms both the high- and low-frequency interval schedulers by sending raw data to the client in a timely manner (on average \SI{13}{\second}). 
The high-frequency fixed scheduler achieves lower latency than its low-frequency counterpart. However, it may require knowledge of when drift is introduced (e.g., for N1, its latency is \SI{7}{\second} by coincidence when there is a match, and it is \SI{415}{\second} for N3 if not). In a real-world deployment, this would not be feasible to achieve. Our method, therefore, provides a practical solution for such scenarios when drift can be experienced at any time. 

\vspace{-1mm}

\subsection{Assessing the Data Communication}
Finally, to evaluate the amount of data transmitted in such a multi-sensor/client setup, we first compared the cumulative data transferred between client 1 and the affected sensor; essentially isolating the affected nodes from the FL system. We then compared the data transmitted between the three schedulers in the 4 clients/32 sensors setup.
By plotting the data communicated between client 1 and the affected sensor, shown in Figure~\ref{fig:IsolatedData}, FLARE performs similarly to the low-frequency scheduler but transmits much less than the high-frequency method.
However, if we consider all the data transmitted in the entire FL system, which includes 4 clients and 32 sensors, the proposed scheduling scheme shows a significant reduction, 
as shown in Figure~\ref{fig:DataTransferred}. 
Furthermore, since both fixed scheduling schemes regularly communicate, they are unsuitable for longer-term deployments where data communication volume is linearly increasing.
The above demonstrate the scalability of FLARE, because, as shown, the amount of data transferred does not change significantly as the length of the experiment increases.

\vspace{-3mm}

\section{Limitations and Future work}
Although our system presents a compelling set of methods for reducing data communication by detecting and reacting to drift in an FL architecture, there are several areas where it could be further optimised. 
Currently, FLARE uses fixed thresholds for detecting drift as well as regulating the frequency of communications. This method demonstrates the potential of our system and the need for similar systems within such FL architectures. 

However, further automated optimisation techniques based on the dataset can also be developed considering various factors such as available data rates.
In the future, we envision developing adaptive thresholding schemes, which will also enable generalisation to other types of data sets \cite{krizhevsky2017imagenet,lin2014microsoft}. One potential method for an adaptive threshold implementation would be to use an observation window during run-time to monitor the models and set thresholds adaptively. This would allow the system to set appropriate thresholds depending on the state of training. 

In this paper, we mainly evaluated the proposed framework when exposed to \textit{abrupt} type of drift. For dealing with other types of drift, such as gradual or incremental \cite{10060415,10060554}, further experiments will be required either using existing datasets that present such behaviours or synthetically generating drift on existing datasets. The proposed system may also require additional research to be able to optimally detect these types of drifts, however, it must still be able to perform in a lightweight manner for deployment in resource-constrained settings.


\section{Conclusion}
In this paper, we proposed detection and mitigation algorithms for distributed learning and deployment systems when they are exposed to dynamic environments, and as such, are subject to concept drift. The main aim of this work was to both detect and react to these changes in an optimised, scalable and timely manner. The proposed methods not only help such a system recover from performance deterioration when exposed to data distribution changes but does it with minimal data communication. We conducted an extensive evaluation of our proposed solution, FLARE, under FL deployments of varying scales, as well as benchmarked against the baseline fixed scheduling methods by comparing accuracy, drift detection latency and data communication volume. When compared against fixed interval schedulers, our proposed solution is able to achieve similar levels of accuracy whilst keeping data transfer to a minimum. FLARE, also has a lower detection latency compared to fixed interval scheduling schemes.

\bibliographystyle{IEEEtran}
\bibliography{References}
\end{document}